\documentclass{ecai2014}
\usepackage{times}
\usepackage{graphicx}
\usepackage{latexsym}
\usepackage{amssymb,amsmath}
\usepackage{algorithm2e}

\newtheorem{definition}{Definition}

\begin{document}

\title{Statistical Constraints}

\author{Roberto Rossi\institute{Business School, University of Edinburgh, Edinburgh, United Kingdom, email: roberto.rossi@ed.ac.uk}
 \and Steven Prestwich\institute{Insight Centre for Data Analytics, University College Cork, Cork, Ireland, email: s.prestwich@cs.ucc.ie} 
 \and S. Armagan Tarim\institute{Institute of Population Studies, Hacettepe University, Ankara, Turkey, email: armagan.tarim@hacettepe.edu.tr} }

\maketitle
\bibliographystyle{ecai2014}

\begin{abstract}
We introduce statistical constraints, a declarative modelling tool that links statistics and constraint programming. We discuss two statistical constraints and some associated filtering algorithms. Finally, we illustrate applications to standard problems encountered in statistics and to a novel inspection scheduling problem in which the aim is to find inspection plans with desirable statistical properties.
\end{abstract}

\section{INTRODUCTION}

Informally speaking, a statistical constraint exploits statistical inference to determine what assignments satisfy a given statistical property at a prescribed significance level. For instance, a statistical constraint may be used to determine, for a given distribution, what values for one or more of its parameters, e.g. the mean, are consistent with a given set of samples. Alternatively, it may be used to determine what sets of samples are compatible with one or more hypothetical distributions. In this work, we introduce the first two examples of statistical constraints embedding two well-known statistical tests: the $t$-test and the Kolmogorov-Smirnov test. Filtering algorithms enforcing bound consistency are discussed for some of the statistical constraints presented. Furthermore, we discuss applications spanning from standard problems encountered in statistics to a novel inspection scheduling problem in which the aim is to find inspection plans featuring desirable statistical properties. 

\section{FORMAL BACKGROUND}

In this section we introduce the relevant formal background. 

\subsection{Statistical inference}\label{sec:inference}


A probability space, as introduced in \cite{kolmogorov1960foundations}, is a mathematical tool that aims at modelling a real-world experiment consisting of outcomes that occur randomly. As such it is described by a triple $(\Omega,\mathcal{F}, \mathcal{P})$, where $\Omega$ denotes the sample space --- i.e. the set of all possible outcomes of the experiment; $\mathcal{F}$ denotes the sigma-algebra on $\Omega$ --- i.e. the set of all possible events on the sample space, where an event is a set that includes zero or more outcomes; and $\mathcal{P}$ denotes the probability measure --- i.e. a function $\mathcal{P}:\mathcal{F}\rightarrow[0,1]$ returning the probability of each possible event. A random variable $\omega$ is an $\mathcal{F}$-measurable function $\omega:\Omega \rightarrow \mathbb{R}$ defined on a probability space $(\Omega,\mathcal{F}, \mathcal{P})$  mapping its sample space to the set of all real numbers. Given $\omega$, we can ask questions such as ``what is the probability that $\omega$ is less or equal to element $s\in\mathbb{R}$.'' This is the probability of event $\{o: \omega(o) \leq s \}\in\mathcal{F}$, which is often written as $F_{\omega}(s)=\Pr(\omega\leq s)$, where $F_{\omega}(s)$ is the cumulative distribution function (CDF) of $\omega$. A multivariate random variable is a random vector $(\omega_1,\ldots,\omega_n)^{T}$, where ${}^{T}$ denotes the ``transpose'' operator. If $\omega_1,\ldots,\omega_n$ are independent and identically distributed (iid) random variables, the random vector may be used to represent an experiment repeated $n$ times, i.e. a sample, where each replica $i$ generates a random variate $\omega'_i$ and the outcome of the experiment is vector $(\omega'_1,\ldots,\omega'_n)^{T}$. 

Consider a multivariate random variable 
defined on probability space $(\Omega,\mathcal{F}, \mathcal{P})$ and let $\mathcal{D}$ be a set of possible CDFs on the sample space $\Omega$. 
In what follows, we adopt the following definition of a statistical model \cite{McCullagh2002}.
\begin{definition}
A statistical model is a pair $\langle \mathcal{D}, \Omega \rangle$.
\end{definition}

Let $\mathbb{D}$ denote the set of all possible CDFs on $\Omega$. Consider a finite-dimensional parameter set $\Theta$ together with a function $g:\Theta\rightarrow \mathbb{D}$, which assigns to each parameter point $\theta\in\Theta$ a CDF $F_{\theta}$ on $\Omega$. 
\begin{definition}
A parametric statistical model is a triple $\langle \Theta, g, \Omega \rangle$.
\end{definition}
\begin{definition}
A non-parametric statistical model is a pair $\langle \mathbb{D}, \Omega \rangle$.
\end{definition}
Note that there are also semi-parametric models, which however for the sake of brevity we do not cover in the following discussion.

Consider now the outcome $o\in \Omega$ of an experiment. Statistics operates under the assumption that there is a distinct element $d\in \mathcal{D}$ that generates the observed data $o$. The aim of statistical inference is then to determine which element(s) are likely to be the one generating the data. A widely adopted method to carry out statistical inference is hypothesis testing.

In hypothesis testing the statistician selects a significance level $\alpha$ and formulates a null hypothesis, e.g. ``element $d\in\mathcal{D}$ has generated the observed data,'' and an alternative hypothesis, e.g. ``another element in $\mathcal{D}/d$ has generated the observed data.'' Depending on the type of hypothesis formulated, she must then select a suitable statistical test and derive the distribution of the associated test statistic under the null hypothesis. By using this distribution, one determines the probability $p_o$ of obtaining a test statistic at least as extreme as the one associated with outcome $o$, i.e. the ``$p$-value''. If this probability is less than $\alpha$, this means that the observed result is highly unlikely under the null hypothesis, and the statistician should therefore ``reject the null hypothesis.'' Conversely, if this probability is greater or equal to $\alpha$, the evidence collected is insufficient to support a conclusion
against the null hypothesis, hence we say that one ``fails to reject the null hypothesis.'' 

In what follows, we will survey two widely adopted tests \cite{sheskin2003handbook}. A parametric test: the Student's $t$-test \cite{Student1908}; and a non-parameteric one: the Kolmogorov-Smirnov test \cite{KolmogorovAN1933,Smirnov1948}. These two tests are relevant in the context of the following discussion.

\subsubsection{Student's $t$-test}

A $t$-test is any statistical hypothesis test in which the test statistic follows a Student's $t$ distribution if the null hypothesis is supported.

The classic one-sample $t$-test compares the mean of a sample to a specified mean. We consider the null hypothesis $H_0$ that ``the sample is drawn from a random variable with mean $\mu$.'' 
The test statistic is 
\[t=\frac{\bar{x}-\mu}{s/\sqrt{n}}\]
where $\bar{x}$ is the sample mean, $s$ is the sample standard deviation and $n$ is the sample size. Since Student's $t$ distribution is symmetric, $H_0$ is rejected if $\Pr(x>t|H_0)<\alpha/2$ or $\Pr(x<t|H_0)<\alpha/2$ that is
\[\mu< \bar{x}+\frac{s}{\sqrt{n}} T^{-1}_{n-1}(\alpha/2)~~~~\text{or}~~~~\mu> \bar{x}-\frac{s}{\sqrt{n}} T^{-1}_{n-1}(\alpha/2)\]
where $T^{-1}_{n-1}$ is the inverse Student's $t$ distribution with $n-1$ degrees of freedom. The respective single-tailed tests can be used to determine if the sample is drawn from a random variable with mean less (greater) than $\mu$.

The two-sample $t$-test compares means $\mu_1$ and $\mu_2$ of two samples. We consider the case in which sample sizes are different, but variance is assumed to be equal for the two samples. The test statistic is
\[t=\frac{\bar{x}_1-\bar{x}_2}{\sqrt{s^2 \left(\frac{1}{n_1}+\frac{1}{n_2}\right)}}~~~~s^2=\frac{\sum_{i=1}^{n_1}(x^i_1-\bar{x}_1)+\sum_{i=1}^{n_2}(x^i_2-\bar{x}_2)}{n_1+n_2-2}\]
where $\bar{x}_1$ and $\bar{x}_2$ are the sample means of the two samples; $s^2$ is the pooled sample variance; $x^j_i$ denotes the $j$th random variate in sample $i$; $n_1$ and $n_2$ are the sample sizes of the two samples; and $t$ follows a Student's $t$ distribution with $n_1+n_2-2$ degrees of freedom. If our null hypothesis is $\mu_1<\mu_2$, it will be rejected if
\[\mu_1-\mu_2+T^{-1}_{n_1+n_2-2}(\alpha)\sqrt{s^2 \left(\frac{1}{n_1}+\frac{1}{n_2}\right)}\geq0\] 
Null hypothesis such as $\mu_1>\mu_2$, $\mu_1\neq \mu_2$ and $\mu_1 = \mu_2$ are tested in a similar fashion.

Note that a range of other test statistics can be used when different assumptions apply \cite{sheskin2003handbook}, e.g. unequal variance between samples.

\subsubsection{Kolmogorov-Smirnov test}\label{sec:ks-test}

The one-sample Kolmogorov-Smirnov (KS) test is a non-parametric test used to compare a sample with a reference CDF defined on a continuous support under the null hypothesis $H_0$ that the sample is drawn from such reference distribution. 

Consider random variates $s=(\omega'_1,\ldots,\omega'_n)^{T}$ drawn from a sample $(\omega_1,\ldots,\omega_n)^{T}$. The empirical CDF $F_s$ is defined as
\[F_s(x)=\frac{1}{n}\sum_{i=1}^n I(\omega'_i\leq x)\]
where the indicator function $I(\omega'_i\leq x)$ is 1 if $\omega'_i\leq x$ and 0 otherwise. For a target CDF $F$, let 
\[d^+_s=\sqrt{n} \sup_{x\in\mathcal{S}}F_s(x)-F(x)~~~\text{and}~~~d^-_s=\sqrt{n} \sup_{x\in\mathcal{S}}F(x)-F_s(x)\]
the KS statistic is 
\[d_s=\max(d^+_s,d^-_s)\]
where $\sup_{x\in\mathcal{S}}$ is the supremum of the set of distances between the empirical and the target CDFs. Under the null hypothesis, $d_s$ converges to the Kolmogorov distribution. Therefore, the null hypothesis is the rejected if $\Pr\{x>d_s|H_0\}<\alpha$, that is
$1-K(d_s)<\alpha$, where $K(t)$
is the CDF of the Kolmogorov distribution, which can be numerically approximated \cite{Pelz1976,Simard:L'Ecuyer:2010:JSSOBK:v39i11}. 

The single-tailed one-sample KS test can be used to determine if the sample is drawn from a distribution that has first-order stochastic dominance over the reference distribution --- i.e.  $F_{\omega} \leq F(x)$ for all $x\in\mathcal{S}$ and with a strict inequality at some $x$ --- in which case the relevant test statistic is $d^+_s$; or vice-versa, in which case the relevant test statistic is $d^-_s$. 

Note that the inverse Kolmogorov distribution $K_n^{-1}$ for a sample of size $n$ can be employed to set a {\em confidence band} around $F$. Let $d_\alpha=K_n^{-1}(1-\alpha)$, then with probability $1-\alpha$ a band of $\pm d_\alpha$ around $F$ will entirely contain the empirical CDF $F_s$.

The two-sample KS test compares two sets of random variates $s_1$ and $s_2$ of size $n_1$ and $n_2$ under the null hypothesis $H_0$ that the respective samples are drawn from the same distribution. Let 
\[d^+_s= \sqrt{\frac{n_1 n_2}{n_1+n_2}}\sup_{x\in\mathcal{S}}F_{s_1}(x)-F_{s_2}(x)\]
\[d^-_s=\sqrt{\frac{n_1 n_2}{n_1+n_2}}\sup_{x\in\mathcal{S}}F_{s_2}(x)-F_{s_1}(x)\]
the test statistic is 
\[d_s=\max(d^+_s,d^-_s)\]

Finally, also in this case it is possible to perform single-tailed tests using test statistics $d^+_s$ or   $d^-_s$ to determine if one of the samples is drawn from a distribution that stochastically dominates the one from which the other sample is drawn.

\subsection{Constraint programming}\label{sec:cp}

A Constraint Satisfaction Problem (CSP) is a triple $\langle V,C,D \rangle$, where $V$
is a set of decision variables, $D$ is a function mapping each element of $V$ to a domain of
potential values, and $C$ is a set of constraints stating allowed combinations of values for subsets of variables in $V$ \cite{1207782}. A {\em solution} to a CSP is an assignment of variables to values in their respective domains such that all of the constraints are satisfied. The constraints used in constraint programming are of various kinds: e.g. logic constraints, linear constraints, and {\em global constraints} \cite{reg03}. A global constraint captures a relation among a non-fixed number of variables. 
Constraints typically embed dedicated {\em filtering algorithms} able to remove provably infeasible
or suboptimal values from the domains of the decision variables that are constrained and, therefore, to enforce some degree of {\em consistency}, e.g. arc consistency, bound consistency \cite{Choi06} or generalised arc consistency. A constraint is \emph{generalized arc consistent} if and only if, when a variable is assigned any of the values in its domain, there exist compatible values in the domains of all the other variables in the constraint. 
Filtering algorithms are repeatedly called until no more values are pruned. This process is called {\em constraint propagation}.
In addition to constraints and filtering algorithms, constraint solvers also feature a heuristic {\em search engine}, e.g. a backtracking algorithm. During search, the constraint solver explores partial assignments and exploits filtering algorithms in order to proactively prune parts of the search space that cannot lead to a feasible or to an optimal solution.

\section{STATISTICAL CONSTRAINTS}


\begin{definition}
A statistical constraint is a constraint that embeds a parametric or a non-parametric statistical model and a statistical test with significance level $\alpha$ that is used to determine which assignments satisfy the constraint.
\end{definition}

A {\bf parametric statistical constraint} $c$ takes the general form $c(T,g,O,\alpha)$; where $T$ and $O$ are sets of decision variables and $g$ is a function as defined in Section \ref{sec:inference}. Let $T\equiv \{t_1,\ldots,t_{|T|}\}$, then $\Theta=D(t_1)\times\ldots\times D(t_{|T|})$. Furthermore, let $O\equiv\{o_1,\ldots,o_{|O|}\}$, then $\Omega= D(o_1)\times\ldots\times D(o_{|O|})$. An assignment is consistent with respect to $c$ if the statistical test fails to reject the associated null hypothesis, e.g. ``$F_\theta$ generated $o_1,\ldots,o_{|O|}$,'' at significance level $\alpha$.

A {\bf non-parametric statistical constraint} $c$ takes the general form $c(O_1,\ldots,O_k,\alpha)$; where $O_1,\ldots,O_k$ are sets of decision variables. Let $O_i\equiv\{o^i_1,\ldots,o^i_{|O_i|}\}$, then $\Omega=\bigcup_{i=1}^k D(o^i_1)\times\ldots\times D(o^i_{|O_i|})$. An assignment is consistent with respect to $c$ if the statistical test fails to reject the associated null hypothesis, e.g ``$\{o^1_1,\ldots,o^1_{|O_1|}\}$,\ldots,$\{o^k_1,\ldots,o^k_{|O_k|}\}$ are drawn from the same distribution,'' at significance level $\alpha$.

In contrast to classical statistical testing, {\em random variates}, i.e. random variable realisations $(\omega'_1,\ldots,\omega'_n)^{T}$, associated with a sample are modelled as decision variables. The {\em sample}, i.e. the set of random variables $(\omega_1,\ldots,\omega_n)^{T}$ that generated the random variates is not explicitly modelled. This modelling strategy paves the way to a number of novel applications. We now introduce a number of parametric and non-parametric statistical constraints. 

\subsection{Parametric statistical constraints}

In this section we introduce two parametric statistical constraints: the Student's $t$ test constraint and the Kolmogorov-Smirnov constraint.

\subsubsection{Student's $t$ test constraint}

Consider statistical constraint 
\[t\text{-test}^\alpha_{w}(O,m)\]
where $O\equiv\{o_1,\ldots,o_n\}$ is a set of decision variables each of which represents a random variate $\omega'_i$; $m$ is a decision variable representing the mean of the random variable $\omega$ that generated the sample. Parameter $\alpha\in(0,1)$ is the significance level; parameter $w\in\{\leq,\geq,=,\neq\}$ identifies the type of statistical test that should be employed, e.g. ``$\leq$'' refers to a single-tailed Student's $t$-test that determines if the mean of $\omega$ is less than or equal to $m$,``$=$'' refers to a two-tailed Student's $t$-test that determines if the mean of $\omega$ is equal to $m$, etc.
An assignment $\bar{o}_1,\ldots,\bar{o}_n,\bar{m}$ satisfies $t\text{-test}^\alpha_{w}$ if and only if a one-sample Student's $t$-test fails to reject the null hypothesis identified by $w$; e.g. if $w$ is ``$=$'', then the null hypothesis is `` the mean of the random variable that generated $\bar{o}_1,\ldots,\bar{o}_n$ is equal to $\bar{m}$.''

The statistical constraint just presented is a special case of 
\[t\text{-test}^\alpha_{w}(O_1,O_2)\]
in which the set $O_2$ contains a single decision variable, i.e. $m$. However, in general $O_2$ is defined as $O_2\equiv\{o_{n+1},\ldots,o_m\}$. In this case, an assignment $\bar{o}_1,\ldots,\bar{o}_m$ satisfies $t\text{-test}^\alpha_{w}$ if and only if a two-sample Student's $t$-test fails to reject the null hypothesis identified by $w$; e.g. if $w$ is ``$=$'', then the null hypothesis is ``the mean of the random variable originating $\bar{o}_1,\ldots,\bar{o}_n$ is equal to that of the random variable generating $\bar{o}_{n+1},\ldots,\bar{o}_m$.''

Note that $t\text{-test}^\alpha_{=}$ is equivalent to enforcing both $t\text{-test}^{1-(1-\alpha)/2}_{\leq}$ and $t\text{-test}^{1-(1-\alpha)/2}_{\geq}$; and that $t\text{-test}^\alpha_{\neq}$ is the complement of $t\text{-test}^\alpha_{=}$.

We leave the development of effective filtering strategies for $t\text{-test}^\alpha_{\leq}$ and $t\text{-test}^\alpha_{\geq}$, which may be based on a strategy similar to that presented in \cite{citeulike:13171963}, as a future research direction. 
  

\subsubsection{Parametric Kolmogorov-Smirnov constraint}

Consider statistical constraint 
\[\text{KS-test}^\alpha_{w}(O,\text{exponential}(\lambda))\]
where $O\equiv\{o_1,\ldots,o_n\}$ is a set of decision variables each of which represents a random variate $\omega'_i$; $\lambda$ is a decision variable representing the rate of the exponential distribution. Note that $\text{exponential}(\lambda)$ may be, in principle, replaced with any other parameterised distribution. However, due to its relevance in the context of the following discussion, in this section we will limit our attention to the exponential distribution. Once more, parameter $\alpha\in(0,1)$ is the significance level; and parameter $w\in\{\leq,\geq,=,\neq\}$ identifies the type of statistical test that should be employed; e.g. ``$\geq$'' refers to a single-tailed one-sample KS test that determines if the distribution originating the sample has first-order stochastic dominance over $\text{exponential}(\lambda)$; ``$=$'' refers to a two-tailed one-sample KS test that determines if the distribution originating the sample is likely to be $\text{exponential}(\lambda)$, etc.

An assignment $\bar{o}_1,\ldots,\bar{o}_n,\bar{\lambda}$ satisfies $\text{KS-test}^\alpha_{w}$ if and only if a one-sample KS test fails to reject the null hypothesis identified by $w$; e.g. if $w$ is ``$=$'', then the null hypothesis is ``random variates $\bar{o}_1,\ldots,\bar{o}_n$ have been sampled from an $\text{exponential}(\lambda)$.''

In contrast to the $t\text{-test}^\alpha_{w}$ constraint, because of the structure of test statistics $d^+_s$ and $d^-_s$, $\text{KS-test}^\alpha_{w}$ is monotonic --- i.e. it satisfies Definition 9 in \cite{Yuanlin2000} --- and bound consistency can be enforced using standard propagation strategies. In Algorithm \ref{algo:bp_ks_test_one_sample_ge} we present a bound propagation algorithm for parametric $\text{KS-test}^\alpha_{\geq}$ when the target CDF $F_\lambda(x)$ is exponential with rate $\lambda$, i.e. mean $1/\lambda$; $\sup(D(x))$ and $\inf(D(x))$ denote the supremum and the infimum of the domain of decision variable $x$, respectively. Note the KS test at lines \ref{alg:param-ks-test-1} and \ref{alg:param-ks-test-2}.

Propagation for parametric $\text{KS-test}^\alpha_{\leq}$ is based on test statistic $d^-_s$ and follows a similar logic. Also in this case $\text{KS-test}^\alpha_{=}$ is equivalent to enforcing both $\text{KS-test}^{1-(1-\alpha)/2}_{\leq}$ and $\text{KS-test}^{1-(1-\alpha)/2}_{\geq}$; $\text{KS-test}^\alpha_{\neq}$ is the complement of $\text{KS-test}^\alpha_{=}$. 
\begin{algorithm}[h!]
\caption{Bound propagation for parametric $\text{KS-test}^\alpha_{\geq}$\label{algo:bp_ks_test_one_sample_ge}}
\DontPrintSemicolon
\KwIn{Decision variables $o_1,\ldots,o_n,\lambda$, and parameter $\alpha$}
\KwOut{Bound consistent $o_1,\ldots,o_n,\lambda$}
$s \gets \{\omega'_1,\ldots,\omega'_n\}$\;
\For{$i \gets 1$ \textbf{to} $n$}{
	$\omega'_i \gets \inf(D(o_i))$\;	
	\For{$j \gets 1$ \textbf{to} $n$, $j\neq i$}{
		$\omega'_j \gets \sup(D(o_j))$\;
	}
	$\bar{\lambda} \gets \sup(D(\lambda))$\;
	$d^+_s\gets \sqrt{n} \sup_{x\in s}F_s(x)-F_{\bar{\lambda}}(x)$\;
\lnl{alg:param-ks-test-1}
	\While{$1-K(d^+_s)<\alpha$}{
		$D(o_i) \gets D(o_i)/\omega'_i$\;
		$\omega'_i \gets \inf(D(o_i))$\;
		$d^+_s\gets \sqrt{n} \sup_{x\in s}F_s(x)-F_{\bar{\lambda}}(x)$\;
	}
}
$\omega'_n \gets \sup(D(o_n))$\;
$\bar{\lambda} \gets \inf(D(\lambda))$\;
$d^+_s\gets \sqrt{n} \sup_{x\in s}F_s(x)-F_{\bar{\lambda}}(x)$\;
\lnl{alg:param-ks-test-2}
\While{$1-K(d^+_s)<\alpha$}{
	$D(\lambda) \gets D(\lambda)/\bar{\lambda}$\;
	$\bar{\lambda} \gets \inf(D(\lambda))$\;
	$d^+_s\gets \sqrt{n} \sup_{x\in s}F_s(x)-F_{\bar{\lambda}}(x)$\;
}
\end{algorithm}

\subsection{Non-parametric statistical constraint}

In this section we introduce a non-parametric version of the Kolmogorov-Smirnov constraint.

\subsubsection{Non-parametric Kolmogorov-Smirnov constraint}

Consider statistical constraint 
\[\text{KS-test}^\alpha_{w}(O_1,O_2)\]
where $O_1\equiv\{o_1,\ldots,o_n\}$ and $O_2\equiv\{o_{n+1},\ldots,o_m\}$ are sets of decision variables representing random variates; once more, parameter $\alpha\in(0,1)$ is the significance level and parameter $w\in\{\leq,\geq,=,\neq\}$ identifies the type of statistical test that should be employed; e.g. ``$\geq$'' refers to a single-tailed two-sample KS test that determines if the distribution originating sample $O_1$ has first-order stochastic dominance over the distribution originating sample $O_2$; ``$=$'' refers to a two-tailed two-sample KS test that determines if the two samples have been originated by the same distribution, etc.

An assignment $\bar{o}_1,\ldots,\bar{o}_m$ satisfies $\text{KS-test}^\alpha_{w}$ if and only if a two-sample KS test fails to reject the null hypothesis identified by $w$; e.g. if $w$ is ``$=$'', then the null hypothesis is ``random variates $\bar{o}_1,\ldots,\bar{o}_n$ and $\bar{o}_{n+1},\ldots,\bar{o}_m$ have been sampled from the same distribution.''

Also in this case the constraint is monotonic and bound consistency can be enforced using standard propagation strategies. In Algorithm \ref{algo:bp_ks_test_two_sample_ge} we present a bound propagation algorithm for non-parametric $\text{KS-test}^\alpha_{\geq}$. Note the KS test at lines \ref{alg:nonparam-ks-test-1} and \ref{alg:nonparam-ks-test-2}.
\begin{algorithm}[h!]
\caption{Bound propagation for non-parametric $\text{KS-test}^\alpha_{\geq}$\label{algo:bp_ks_test_two_sample_ge}}
\DontPrintSemicolon
\KwIn{Decision variables $o_1,\ldots,o_m$, and parameter $\alpha$}
\KwOut{Bound consistent $o_1,\ldots,o_m$}
$s_1 \gets \{\omega'_1,\ldots,\omega'_n\}$\;
$s_2 \gets \{\omega'_{n+1},\ldots,\omega'_m\}$\;
$n_1 \gets n$\;
$n_2 \gets m-n$\;
\For{$i \gets 1$ \textbf{to} $n$}{
	$\omega'_i \gets \inf(D(o_i))$\;	
	\For{$j \gets 1$ \textbf{to} $n$, $j\neq i$}{
		$\omega'_j \gets \sup(D(o_j))$\;
	}
	
	\For{$j \gets n+1$ \textbf{to} $m$}{
		$\omega'_j \gets \inf(D(o_j))$\;
	}
	
	$d^+_s\gets\sqrt{\frac{n_1 n_2}{n_1+n_2}}\sup_{x\in s_1 \cup s_2}F_{s_1}(x)-F_{s_2}(x)$\;
\lnl{alg:nonparam-ks-test-1}
	\While{$1-K(d^+_s)<\alpha$}{
		$D(o_i) \gets D(o_i)/\omega'_i$\;
		$\omega'_i \gets \inf(D(o_i))$\;
		$d^+_s\gets\sqrt{\frac{n_1 n_2}{n_1+n_2}}\sup_{x\in s_1 \cup s_2}F_{s_1}(x)-F_{s_2}(x)$\;
	}
}

\For{$i \gets n+1$ \textbf{to} $m$}{
	$\omega'_i \gets \sup(D(o_i))$\;	
	\For{$j \gets n+1$ \textbf{to} $m$, $j\neq i$}{
		$\omega'_j \gets \inf(D(o_j))$\;
	}
	
	\For{$j \gets 1$ \textbf{to} $n$}{
		$\omega'_j \gets \sup(D(o_j))$\;
	}
	
	$d^+_s\gets\sqrt{\frac{n_1 n_2}{n_1+n_2}}\sup_{x\in s_1 \cup s_2}F_{s_2}(x)-F_{s_1}(x)$\;
\lnl{alg:nonparam-ks-test-2}
	\While{$1-K(d^+_s)<\alpha$}{
		$D(o_i) \gets D(o_i)/\omega'_i$\;
		$\omega'_i \gets \sup(D(o_i))$\;
		$d^+_s\gets\sqrt{\frac{n_1 n_2}{n_1+n_2}}\sup_{x\in s_1 \cup s_2}F_{s_2}(x)-F_{s_1}(x)$\;
	}
}
\end{algorithm}

Propagation for non-parametric $\text{KS-test}^\alpha_{\leq}$ is based on test statistic $d^-_s$ and follows a similar logic. Also in this case $\text{KS-test}^\alpha_{=}$ is equivalent to enforcing both $\text{KS-test}^{1-(1-\alpha)/2}_{\leq}$ and $\text{KS-test}^{1-(1-\alpha)/2}_{\geq}$; $\text{KS-test}^\alpha_{\neq}$ is the complement of $\text{KS-test}^\alpha_{=}$. 

\section{APPLICATIONS}

In this section we discuss a number of applications for the statistical constraints previously introduced. 

\subsection{Classical problems in statistics}\label{sec:statistics_applications}

In this section we discuss two simple applications in which statistical constraints are employed to solve classical problems in hypothesis testing. The first problem is parametric, while the second is non-parametric.

The first application is a standard $t$-test on the mean of a sample. Given a significance level $\alpha=0.05$ and random variates $\{8, 14, 6, 12, 12, 9, 10, 9, 10, 5\}$ we are interested in finding out the mean of the random variable originating the sample. This task can be accomplished via a CSP such as the one in Fig. \ref{model:t-test_mean}.
\begin{figure}[h!]
\begin{center}
        \framebox{
        \begin{tabular}{ll}
        \mbox{\bf Constraints:}\\
        ~~~~(1)~~$t\text{-test}^\alpha_{=}(O,m)$\\
        \mbox{\bf Decision variables:} \\
        ~~~~$o_1\in\{8\}, o_2\in\{14\}, o_3\in\{6\}, o_4\in\{12\}, o_5\in\{12\},$\\
        ~~~~$o_6\in\{9\}, o_7\in\{10\}, o_8\in\{9\}, o_9\in\{10\}, o_{10}\in\{5\}$\\
        ~~~~$O_1\equiv\{o_1,\ldots,o_{10}\}$\\
        ~~~~$m \in \{0,\ldots,20\}$
        \end{tabular}
        }
    \caption{Determining the likely values of the mean of the random variable that generated random variates $O_1$}
    \label{model:t-test_mean}
\end{center}    
\end{figure}
After propagating constraint (1), the domain of $m$ reduces to $\{8,9,10,11\}$, so with significance level $\alpha=0.05$ we reject the null hypothesis that the true mean is outside this range.
Despite the fact that in this work we do not discuss a filtering strategy for the $t\text{-test}$ constraint, in this specific instance we were able to propagate this constraints due to the fact that all decision variables $o_i$ were ground. In general the domain of these variables may not be a singleton. In the next example we illustrate this case.

\begin{figure}[h!]
\begin{center}
        \framebox{
        \begin{tabular}{ll}
        \mbox{\bf Constraints:}\\
        ~~~~(1)~~$\text{KS-test}^\alpha_{=}(O_1,O_2)$\\
        \mbox{\bf Decision variables:} \\
        ~~~~$o_1\in\{9\}, o_2\in\{10\}, o_3\in\{9\}, o_4\in\{6\}, o_5\in\{11\},$\\
        ~~~~$o_6\in\{8\}, o_7\in\{10\}, o_8\in\{11\}, o_9\in\{14\}, o_{10}\in\{11\},$\\
        ~~~~$o_{11},o_{12}\in\{5\}, o_{13},\ldots,o_{20}\in\{9,10,11\}$\\
        ~~~~$O_1\equiv\{o_1,\ldots,o_{10}\}, O_2\equiv\{o_{11},\ldots,o_{20}\}$
        \end{tabular}
        }
    \caption{Devising sets of random variates that are likely to be generated from the same random variable that generated a reference set of random variates $O_1$}
    \label{model:KS-test}
\end{center}    
\end{figure}
\begin{figure}[t]
\centerline{\includegraphics[type=eps,ext=.eps,read=.eps,width=1\columnwidth]{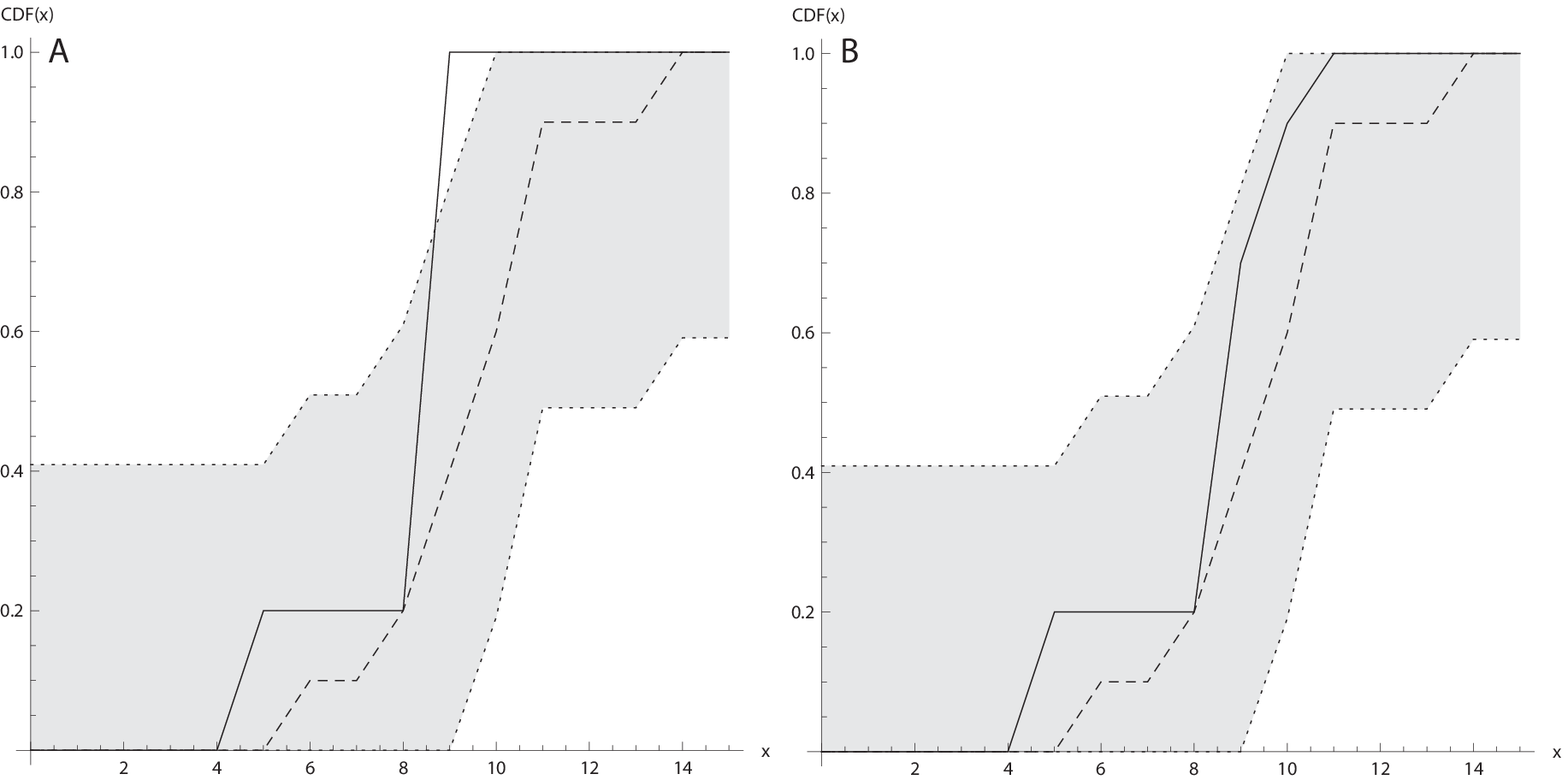}}
\caption{Empirical CDFs of (A) an infeasible and of (B) a feasible set of random variates $O_2$ for the CSP in Fig. \ref{model:KS-test}; these are $\{5, 5, 9, 9, 9, 9, 9, 9, 9, 9\}$ and $\{5, 5, 9, 9, 9, 9, 9, 10, 10, 11\}$, respectively.}
\label{fig:KS-test}
\end{figure}

Consider the CSP in Fig. \ref{model:KS-test}. Decision variables in $O_1$ are ground, this choice is made for illustrative purposes --- in general variables in $O_1$ may feature larger domains. Decision variables in $O_2$ feature non-singleton domains. The problem is basically that of finding a subset of the cartesian product $D(o_{11})\times \ldots\times D(o_{20})$ such that for all elements in this set a KS test fails to reject at significance $\alpha=0.05$ the null hypothesis that $O_2$ does not originate from the same random variable that generated $O_1$. Since 8 variables have domains featuring 3 elements there are $6561$ possible sets of random variates. By finding all solutions to the above CSP we verified that there are 365 sets of random variates for which the null hypothesis is rejected at significance level $\alpha$. In Fig. \ref{fig:KS-test}A we show the empirical CDF (black continuous line) of an infeasible set of random variates; while in Fig. \ref{fig:KS-test}B we show that of a feasible set of random variates. The dashed line is the empirical CDF of the reference set of random variates $O_1$, the grey area is the confidence band around this empirical CDF, obtained as discussed in Section \ref{sec:ks-test}. Recall that, with probability less than $\alpha$, the random variable that originates $O_1$ generates an empirical CDF not fully contained within this area. For clarity, we interpolated the two original stepwise empirical CDFs.

In this latter example we addressed the problem of finding a set of random variates that meets certain statistical criteria. We next demonstrate how similar models can be employed to design inspection plans. 

\subsection{Inspection scheduling}\label{sec:inspection}

We introduce the following inspection scheduling problem. There are 10 units to be inspected 25 times each over a planing horizon comprising 365 days. An inspection lasts 1 day and requires 1 inspector. There are 5 inspectors in total that can carry out inspections at any given day. The average rate of inspection $\lambda$ should be 1 inspection every 5 days. However, there is a further requirement that inter arrival times between subsequent inspections at the same unit of inspection should be {\em approximately exponentially distributed} --- in particular, if the null hypothesis that intervals between inspections follows an exponential($\lambda$) is rejected at significance level $\alpha=0.1$ then the associated plan should be classified as infeasible. This in order to mimic a ``memoryless'' inspection plan, so that the probability of facing an inspection at any given point in time is independent of the number of past inspections; which is clearly a desirable property for an inspection plan. 

\begin{figure}[h!]
\begin{center}
        \framebox{
        \begin{tabular}{ll}
        \mbox{\bf Parameters:}\\
        \begin{tabular}{ll}
        ~~~~$U=10$&Units to be inspected\\  
        ~~~~$I=25$&Inspections per unit\\
        ~~~~$H=365$&Periods in the planning horizon\\
        ~~~~$D=1$&Duration of an inspection\\  
        ~~~~$M=36$&Max interval between two inspections\\  
        ~~~~$C=1$&Inspectors required for an inspection\\  
        ~~~~$m=5$&Inspectors available\\  
        ~~~~$\lambda=1/5$&Inspection rate
        \end{tabular}
        \\
        \mbox{\bf Constraints:}\\
        ~~~~(1)~~$\mathrm{cumulative}(s,e,t,c,m)$ \\
        ~~~~{\bf for all} $u \in 1,\ldots,U$\\
        ~~~~~~~(2)~~$\text{KS-test}^\alpha_{=}(O_u,\text{exponential}(\lambda))$\\
        ~~~~~~~(3)~~$e_{uI} \geq H-M$\\
        ~~~~{\bf for all} $u \in 1,\ldots,U$ {\bf and} $j \in 2,\ldots,I$\\
        ~~~~~~~(4)~~$i_{u,j-1}=s_{uI+j}-s_{uI+j-1}-1$\\
        ~~~~~~~(5)~~$s_{uI+j}\geq s_{uI+j-1}$\\
        \mbox{\bf Decision variables:} \\
        \begin{tabular}{ll}
        ~~~~$s_k \in \{1,\ldots,H\}$,&$\forall k \in 1,\ldots,I\cdot U$\\
        ~~~~$e_k \in \{1,\ldots,H\}$,&$\forall k \in 1,\ldots,I\cdot U$\\
        ~~~~$t_k \gets D$,&$\forall k \in 1,\ldots,I\cdot U$\\
        ~~~~$c_k \gets C$,&$\forall k \in 1,\ldots,I\cdot U$\\
        ~~~~$i_{u,j-1} \in \{0,\ldots,M\}$,&$\forall u \in 1,\ldots,U$ and\\
        &$\forall j \in 2,\ldots,I$\\
        ~~~~$O_u \equiv \{i_{u,1},\ldots, i_{u,I-1}\}$,&$\forall u \in 1,\ldots,U$
        \end{tabular}        
        \end{tabular}
        }
    \caption{Inspection scheduling}
    \label{model:inspection_CSP}
\end{center}    
\end{figure}
This problem can be modelled via the cumulative constraint \cite{citeulike:4257618} as shown in Fig. \ref{model:inspection_CSP}, where $s_k$, $e_k$ and $t_k$ are the start time, end time and duration of inspection $k$; finally $c_k$ is the number of inspectors required to carry out an inspection. The memoryless property of the inspection plan can be ensured by introducing decision variables $i_{u,j-1}$ that model the interval between inspection $j$ and inspection $j-1$ at unit of inspection $u$ (constraint 4). Then, for each unit of inspection $u$ we enforce a statistical constraint $\text{KS-test}^\alpha_{=}(O_u,\text{exponential}(\lambda))$, where $O_u$ is the list of intervals between inspections at unit of inspection $u$. Note that it is possible to introduce side constraints: in this case we force the interval between two consecutive inspections to be less or equal to $M$ days and we make sure that the last inspection is carried out during the last month of the year (constraint 3).

\begin{figure}[h!]
\centerline{\includegraphics[type=eps,ext=.eps,read=.eps,width=0.95\columnwidth]{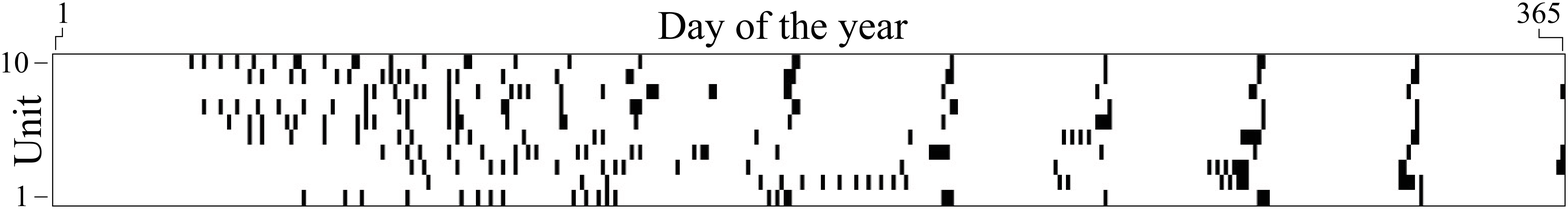}}
\caption{Inspection plan; black marks denote inspections.} 
\label{fig:inspection_plan}
\end{figure}
\begin{figure}[h!]
\centerline{\includegraphics[type=eps,ext=.eps,read=.eps,width=1\columnwidth]{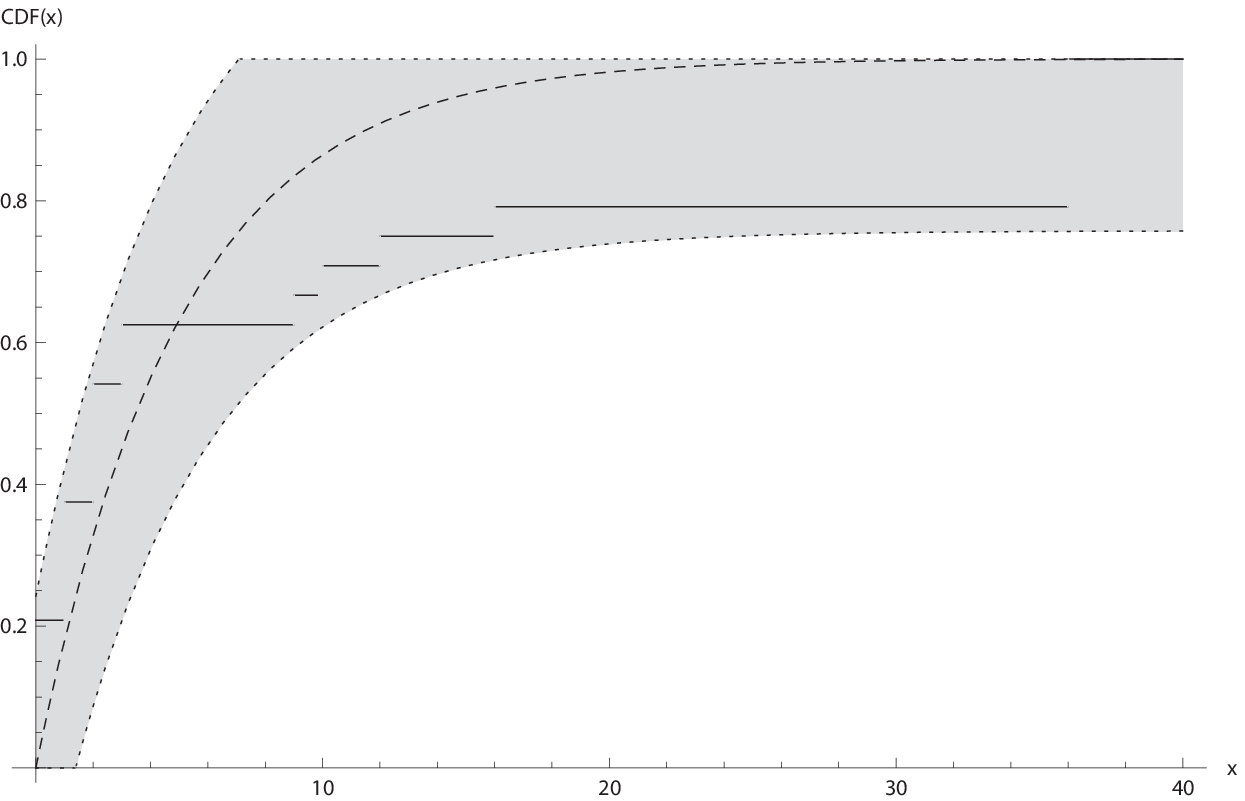}}
\caption{Empirical CDF of intervals (in days) between inspections for unit of assessment 1} 
\label{fig:inspection_stat_constraint}
\end{figure}

In Fig. \ref{fig:inspection_plan} we illustrate a feasible inspection plan for the 10 units of assessment over a 365 days horizon. In Fig. \ref{fig:inspection_stat_constraint} we show that the inspection plan for unit of assessment 1 --- first from the bottom in Fig. \ref{fig:inspection_plan} ---  satisfies the statistical constraint. In fact, the empirical CDF of the intervals between inspections (black stepwise function) is fully contained within the confidence bands of an exponential($\lambda$) distribution (dashed function) at significance level $\alpha$.

\vspace{-3em}

\subsection{Further application areas}\label{sec:future}

The techniques discussed in this work may be used in the context of classical problems encountered in statistics \cite{sheskin2003handbook}, e.g. regression analysis, distribution fitting, etc. In other words, one may look for solutions to a CSP that fit a given set of random variates or distributions. In addition, as seen in the case of inspection scheduling, statistical constraints may be used to address the inverse problem of designing sampling plans that feature specific statistical properties; such analysis may be applied in the context of design of experiments \cite{citeulike:13052847} or quality management \cite{citeulike:8631744}. Further applications may be devised in the context of supply chain coordination. For instance, one may identify replenishment plans featuring desirable statistical properties, e.g. obtain a production schedule in which the ordering process, while meeting other technical constraints, mimics a given stochastic process, e.g. Poisson($\lambda$); this information may then be passed upstream to suppliers to ensure coordination without committing to a replenishment plan fixed a priori or to a specific replenishment policy. 

\section{RELATED WORKS}
The techniques here presented generalise the discussion in \cite{citeulike:9453442}, in which statistical inference is applied in the context of stochastic constraint satisfaction to identify approximate solutions featuring given statistical properties. However, stochastic constraint programming \cite{DBLP:conf/ecai/Walsh02} works with decision and random variables over a set of decision stages; random variable distributions are assumed to be known. Statistical constraints instead operate under the assumption that distribution of random variables is only partially specified (parametric statistical constraints) or not specified at all (non-parametric statistical constraints); furthermore, statistical constraints do not model explicitly random variables, they model instead sets of random variates as decision variables. Finally, a related work is \cite{citeulike:13171963} in which the authors introduce the SPREAD constraint. Like statistical constraints SPREAD ensures that a collection of values exhibits given statistical properties, e.g. mean, variance or median, but its semantic does not feature a significance level. 


\section{CONCLUSION}
Statistical constraints represent a bridge that links statistical inference and constraint programming for the first time in the literature. The declarative nature of constraint programming offers a unique opportunity to exploit statistical inference in order to identify sets of assignments featuring specific statistical properties. Beside introducing the first two examples of statistical constraints, this work discusses filtering algorithms that enforce bound consistency for some of the constraints presented; as well as applications spanning from standard problems encountered in statistics to a novel inspection scheduling problem in which the aim is to find inspection plans featuring desirable statistical properties. 

\vspace{1em}
\noindent
{\bf Acknowledgements}: 
We would like to thank the anonymous reviewers for their valuable suggestions. R. Rossi is supported by the University of Edinburgh CHSS Challenge Investment Fund. S.A. Tarim is supported by the Scientific and Technological Research Council of Turkey (TUBITAK) Project No: 110M500 and by Hacettepe University-BAB. This publication has emanated from research supported in part by a research grant from Science Foundation Ireland (SFI) under Grant Number SFI/12/RC/2289.

\bibliography{ecai2014}
\end{document}